# From Fuzzy Speech to Medical Insight: Benchmarking LLMs on Noisy Patient Narratives

Eden Mama
*Department of Digital Medical Technologies*
Holon Institute of Technology

Liel Sheri
*Department of Digital Medical Technologies*
Holon Institute of Technology

Yehudit Aperstein
*Intelligent Systems,*
*Afeka Academic College of Engineering*
Tel Aviv Israel

Alexander Apartsin
*School of Computer Science,*
*Faculty of Sciences*
Holon Institute of Technology

*Abstract*—The widespread adoption of large language models (LLMs) in healthcare raises critical questions about their ability to interpret patient-generated narratives, which are often informal, ambiguous, and noisy. Existing benchmarks typically rely on clean, structured clinical text, offering limited insight into model performance under realistic conditions. In this work, we present a novel synthetic dataset designed to simulate patient self-descriptions characterized by varying levels of linguistic noise, fuzzy language, and layperson terminology.

Our dataset comprises clinically consistent scenarios annotated with ground-truth diagnoses, spanning a spectrum of communication clarity to reflect diverse real-world reporting styles. Using this benchmark, we fine-tune and evaluate several state-of-the-art models (LLMs), including BERT-based and encoder-decoder T5 models. To support reproducibility and future research, we release the Noisy Diagnostic Benchmark (NDB), a structured dataset of noisy, synthetic patient descriptions designed to stress-test and compare the diagnostic capabilities of large language models (LLMs) under realistic linguistic conditions.

We made the benchmark available for the community: https://github.com/lielsheri/PatientSignal

## I. INTRODUCTION

Large language models (LLMs) such as GPT-4, Med-PaLM 2, and Claude 3 have demonstrated impressive capabilities across a range of natural language processing (NLP) tasks, including question answering, summarization, and even medical diagnosis. However, their performance is typically benchmarked on well-structured, noise-free datasets drawn from clinical notes, exam-style questions, or curated case reports. In contrast, real-world patient communication, especially in initial self-reported symptom descriptions, is rarely so clean or precise. Patients often use vague terminology, layman's expressions, emotional cues, or colloquial phrasing to describe their health concerns. These narratives may include typographical errors, run-on sentences, or incomplete descriptions, making them challenging for both human clinicians and AI systems to interpret reliably.

This mismatch between the training and evaluation conditions of LLMs and the reality of patient communication represents a critical blind spot in current clinical NLP research. While LLMs are increasingly being explored for virtual triage, digital symptom checking, and AI-assisted consultation, their ability to extract and reason over noisy, ambiguous patient inputs remains underexplored and largely unquantified.

In this work, we address this gap by constructing a synthetic benchmark specifically designed to simulate noisy, real-world patient self-descriptions. Each entry in our dataset represents a plausible scenario in which a patient attempts to articulate symptoms using informal, imprecise, or partially incorrect language. The dataset encompasses a range of diagnostic categories and is annotated with ground-truth diagnoses, allowing for quantitative comparison across models. To simulate a realistic communication spectrum, we systematically introduce varying levels and types of noise, including lexical, syntactic, and semantic distortion, while maintaining clinical plausibility.

Our key contributions are as follows:
- **We design and release the Noisy Diagnostic Benchmark (NDB)**, a synthetic dataset of realistic patient self-descriptions with varying noise levels and annotated diagnoses.
- **We fine-tune and evaluate several leading large language models (LLMs)**, including a fine-tuned generic BERT model and a specialized Clinical BERT, along with a fine-tuned generative flan-T5 model, on their ability to infer diagnoses from these noisy narratives.

This work introduces a critical diagnostic stress test for large language models (LLMs) and lays the groundwork for future research into robust, patient-facing clinical AI systems.

## II. LITERATURE REVIEW

### A. LLMs for Diagnostic Reasoning from Text

Large language models (LLMs) have rapidly advanced over the past five years, transforming natural language processing in healthcare. Early transformer-based models, such as BERT, were adapted for clinical text (Alsentzer et al., 2019). However, these models had hundreds of millions of parameters, which are relatively small by today's standards. Recent large language models (LLMs) with billions of parameters, such as GPT-3 and GPT-4, demonstrate impressive capabilities in understanding and generating medical language (Singhal et al., 2023). These models can interpret free-form text ranging from patient descriptions of symptoms to clinician-authored notes. According to Lee, Bubeck, and Petro (2023), the general-purpose GPT-4 chatbot "could affect the practice of medicine" by assisting in clinical reasoning and summarization tasks. This review examines how large language models (LLMs) are being utilized to extract diagnostic insights from unstructured text, how they handle noisy or colloquial patient language, and what datasets of patient-generated text are available for research purposes.

Recent studies show that LLMs can interpret clinical text to produce differential diagnoses or answer diagnostic questions. For example, *Med-PaLM*, an LLM tuned for medical tasks, achieved 67.6% accuracy on USMLE board-style questions, outperforming previous models by over 17%, though it still underperformed practicing clinicians (Singhal et al., 2023). Similarly, Nori et al. (2023) evaluated GPT-4 on medical challenge problems and found it "easily tops prior leading results for medical benchmarks", including expert-level scores on many knowledge tests. Beyond exams, LLMs have been tested on clinical vignettes and live cases. Goh et al. (2024) conducted a randomized trial in which physicians solved cases with or without the assistance of GPT-4. They found no significant overall improvement in diagnostic accuracy with the AI assistant; however, specific components of reasoning, such as contextual decision-making, showed improvement with the aid of LLM. According to Goh and colleagues, providing GPT-4 as a diagnostic aid "did not significantly improve clinical reasoning compared to conventional resources, highlighting that current LLMs are not a panacea. Nonetheless, LLM alone, as used in that study, performed on par with physicians in identifying correct diagnoses, underscoring the potential of these models when used appropriately.

LLMs have also been applied to free-form patient descriptions and conversation transcripts to facilitate the generation of diagnoses. Chen et al. (2025) introduced a multi-agent language model (LLM) approach inspired by how medical teams discuss cases. In their framework, multiple GPT-based "doctor agents" debate a case, and a supervisor agent synthesizes their opinions (Chen et al., 2025). This Multi-Agent Conversation (MAC) model, utilizing GPT-4 as its base, achieved higher diagnostic accuracy than a single Large Language Model (LLM) on a set of 302 rare disease cases in both initial and follow-up consultations (Chen et al., 2025). For instance, with four collaborating agent instances, the most likely diagnosis was correct 34% of the time during the first consultation, compared to 24% when using GPT-3.5 alone. Such results suggest that prompting LLMs to *"think"* in a dialogue emulating specialist consultations can enhance their diagnostic reasoning performance. Another notable system is AMIE (Articulate Medical Intelligence Explorer), an LLM optimized for differential diagnosis (McDuff et al., 2025). In a Nature study, McDuff et al. demonstrated that AMIE, when presented with a case history, could produce a correct diagnosis in its top-10 differential list 59% of the time – significantly outperforming unaided clinicians (33.6% on the same cases). AMIE's suggestions surpassed physicians' diagnostic accuracy under various conditions. According to the authors, these LLM-driven interfaces offer *"new opportunities to assist and automate"* parts of the diagnostic process by interacting with clinicians through chat-based queries (McDuff et al., 2025).

Apart from generating differentials, LLMs have been used to extract key diagnostic information from clinical narratives. Yang et al. (2022) developed GatorTron, an 8.9-billion-parameter clinical large language model (LLM) trained on over 90 billion words of electronic health record text, to assist in interpreting notes. GatorTron improved clinical information extraction tasks, such as identifying medical problems and inferring relations, compared to smaller models, with nearly 10% higher accuracy in some reasoning tasks (nature.com). This demonstrates how scaling up model size and training data enhances the ability to capture subtle clinical clues in text. Similarly, Huang et al. (2024) evaluated ChatGPT for parsing free-text pathology reports into structured data. In their study, GPT-3.5 correctly extracted cancer pathology findings with 89% accuracy, outperforming traditional rule-based methods (Huang et al., 2024). The LLM reliably translated narrative reports into tumor classifications and margin status, predominantly when guided by carefully designed prompts. According to Huang and colleagues, this showcases the feasibility of using general LLMs *"to process large volumes of clinical notes for structured information extraction"* without extensive task-specific training.

Notably, LLMs are also being utilized to summarize and organize patient information, thereby supporting diagnosis. For example, in emergency settings, an LLM might summarize a triage nurse's free-text note and suggest possible diagnoses. Researchers are examining LLM-generated "problem lists" derived from admission notes to determine if they align with clinicians' assessments (Laï et al., 2023). Early results suggest that LLMs can draft reasonable problem summaries but may overlook nuanced context and encounter reliability issues (Laï et al., 2023). As Lee, Bubeck, and Petro (2023) caution, a significant risk associated with GPT-based assistants is their tendency to occasionally produce incorrect or fabricated facts (hallucinations) while appearing confident and credible. This is especially concerning in diagnosis, where an AI might *appear* confident in a suggestion that lacks support. Efforts are underway to mitigate these issues by integrating retrieval (e.g., querying medical databases) or by requiring large language models (LLMs) to provide uncertainty estimates (Zhou et al., 2023). Despite these challenges, there is optimism that, with proper oversight, LLMs can reduce clinician workload by sifting through notes and highlighting relevant findings (Ayers et al., 2023). For instance, in a study on an online forum, ChatGPT's answers to patient questions were preferred over those of physicians 78% of the time and were rated higher in both quality and empathy (Ayers et al., 2023). Such results suggest that LLMs could triage routine patient queries or draft initial responses, enabling physicians to focus on confirming diagnoses and addressing complex cases.

A key challenge in using patient-generated text is that it often contains misspellings, slang, and ambiguous descriptions of symptoms. LLMs trained on large, diverse corpora have shown surprising robustness to such noise. For example, a BERT-based model fine-tuned for phenotype extraction, PhenoBCBERT, was able to recognize misspelled clinical terms in text that stymied earlier tools (Yang et al., 2023). Yang et al. note that PhenoBCBERT remained accurate even *"after replacing random letters"* in phenotype words, successfully identifying the intended medical concept. This robustness is attributed to the model's sub-word tokenization and context awareness, which allow it to infer meaning from surrounding words. Large generative models, such as ChatGPT, similarly handle many colloquial expressions. They often know, for instance, that "heart attack" and "MI" refer to the same condition or that "sugar is high" likely means elevated blood glucose. According to Singhal et al. (2023), instruction-tuned LLMs encode substantial clinical knowledge, including layman terminology, which helps them interpret consumer health questions. Indeed, ChatGPT was able to infer correct differential diagnoses from informal descriptions of symptoms on social media in several case studies (Ayers et al., 2023). Clinicians have observed it translating a patient's vague complaint, such as "feels like pins and needles in my arms," into the medical concept of paraesthesia and then suggesting relevant causes. This translation ability stems from LLM pretraining on enormous text corpora that include forums, Q&A sites, and health articles written in colloquial language. As a result, the model has implicitly learned mappings between everyday language and medical jargon.

However, LLMs are not immune to misunderstanding atypical slang or rare dialectal phrases. They may also be led astray by inconsistencies or irrelevant details in a patient's narrative. A recent analysis found that ChatGPT's diagnostic reasoning may be biased by the way information is presented in the prompt (Huang et al., 2023). If extraneous or misleading context was embedded in the patient history, the model sometimes gave undue weight to it. For example, if a prompt subtly suggested a psychiatric origin (e.g., mentioning the patient is "anxious about her symptoms"), the AI might lean towards psychosomatic explanations. Khatri et al. (2023) reported that ChatGPT was *"sensitive to bias when the biasing information is part of the patient's disease history"*, even if the same situational context alone would not bias the model. This suggests that LLMs still lack a comprehensive understanding and may reflect any assumptions present in the input. To address such issues, researchers are developing techniques such as chain-of-thought prompting, where the model is asked to explicitly reason through inconsistencies, and multi-step cross-checking, where the model re-evaluates its conclusions after being shown its draft reasoning (Chen et al., 2025). These methods aim to have the AI catch its potential mistakes (such as contradicting evidence in the text) before finalizing a diagnosis. Another approach is to incorporate domain-specific constraints or knowledge bases. For instance, an LLM can be guided by a list of verified symptom-disease mappings, ensuring it does not hallucinate a link that is not medically supported (Zhou et al., 2023). Preliminary work on *"trustworthy AI"* for healthcare suggests that grounding LLM outputs in reliable references can reduce factual errors and make the model more robust to noisy input (Zhou et al., 2023). In practice, this might mean an AI clinician that, when uncertain, asks the patient for clarification or recommends standard diagnostic tests rather than making a guess. As Chen et al. (2025) showed with their multi-agent system, having the LLM simulate a consultation, effectively asking itself follow-up questions, can resolve ambiguity in the patient's story and improve accuracy. In summary, while LLMs handle slang and typos more effectively than prior NLP systems, careful prompt design and augmentation with medical knowledge are still necessary to navigate the complexity and occasional messiness of patient-generated text.

B.    *Publicly Available Datasets of Patient Text for LLMs*

Research progress in this area has been aided by the creation of datasets that capture free-form patient conversations and narratives. A variety of **publicly available corpora** now exist for training or evaluating LLMs on patient-generated text:

**MedDialog Dataset (Zeng et al., 2020):** An extensive collection of medical dialogues from an online health forum, containing 260,000 English conversations between patients and doctors (and an even larger Chinese subset). Patients ask questions or describe symptoms in free text, and physicians respond. This dataset encompasses 96 medical specialties, offering a rich resource for modelling doctor-patient interactions. Zeng et al. (2020) demonstrated that dialogue models trained on MedDialog can generate *"clinically correct and doctor-like"* responses. These models have been used to fine-tune large language models (LLMs) for medical question-and-answer (Q&A) and conversational diagnosis tasks.

**Simulated Medical Interviews (Fareez et al., 2022):** A curated dataset of simulated patient-physician conversations following Objective Structured Clinical Examination (OSCE) scenarios (Fareez et al., 2022). It includes transcripts of 167 respiratory case interviews, as well as other specialties, where medical professionals acted out patient and doctor roles. These conversations are lengthy, averaging ~95 turns, and cover history-taking, symptoms, and follow-up questions in detail (Parikh et al., 2022). The data was manually corrected and annotated. This dataset has been utilized to train and evaluate dialogue systems for history-taking and to develop annotation schemes for symptoms within the MediTOD project (Tandon et al., 2024). Because it is synthetically created with expert input, it mitigates privacy concerns while providing a realistic structure for medical dialogue.

**MIMIC-III and MIMIC-IV Clinical Notes (Johnson et al., 2023):** Although not patient conversations, the MIMIC databases serve as a cornerstone for patient narrative data. MIMIC-IV, released in 2023, contains de-identified electronic health record data from ICU admissions, including free-text clinician notes such as history of present illness, discharge summaries, and progress notes (Johnson et al., 2023). These notes often incorporate or paraphrase the patient's own words and cover a wide range of conditions. MIMIC is publicly accessible to researchers under a data use agreement and has been widely used to train and test clinical NLP models. For example, ClinicalBERT was initialized on MIMIC notes to imbue BERT with medical context (Alsentzer et al., 2019). For LLM research, MIMIC notes can be used to have models predict diagnoses or outcomes from the unstructured text, or to evaluate how well an LLM summarizes a patient's hospital course. One limitation is that the text is authored in technical language by clinicians rather than by patients, but it remains invaluable for developing diagnostic models from narrative data.

**Ambient Clinical Intelligence Benchmark (ACI-Bench, Yim et al., 2023):** ACI-Bench is a recent open dataset designed to facilitate training AI "scribes" and diagnostic assistants on doctor-patient dialogue. Unveiled in 2023, it consists of synthetic clinic visit dialogues paired with their corresponding physician-written notes (Yim et al., 2023). The dialogues were generated to reflect primary care encounters, and the notes summarize key findings and the diagnosis. ACI-Bench comprises several hundred encounters and is the largest publicly available dataset for the task of automatic clinical note generation. While its primary use is training models to produce accurate visit summaries, which reduces doctors' paperwork, it is also helpful for diagnosis-focused research. An LLM can be evaluated on how well it transforms a patient-doctor conversation into a structured note with the correct diagnosis and plan. The availability of both the transcript and the gold-standard note makes it ideal for end-to-end evaluation of LLM understanding.

**Synthetic and Augmented Dialogue Datasets:** In addition to real or simulated data, researchers have created synthetic patient dialogues to augment training. Chintagunta et al. (2021) developed a method using GPT-3 to generate artificial medical dialogues, with a focus on ensuring that key medical facts are present. By prompting GPT-3 with medically relevant information (e.g., symptom lists), they generated dialogues and summaries that could supplement smaller real datasets. Such synthetic data has been used to train dialogue models that perform robustly on medical reasoning tasks (Chintagunta et al., 2021). Likewise, the MEDIQA-Chat 2023 challenge released some annotated doctor-patient conversation snippets for note-generation tasks (Ben Abacha et al., 2023). Although limited in size, these shared-task datasets offer valuable testbeds for evaluating LLMs on realistic conversations and assessing the accuracy of extracted information.

**Consumer Health Question Datasets:** Although not dialogues per se, these corpora contain free-text patient questions and descriptions that are useful for model evaluation. One example is the HealthSearchQA dataset introduced by Singhal et al. (2023), which comprises medical questions that users typed into search engines. Each question is free-form, often containing lay terminology and noise, with an associated answer. Similarly, the AskDocs repository, used by Ayers et al. (2023), contains thousands of patient-posted questions on Reddit, along with verified physician answers. These datasets enable testing LLMs on understanding messy, layperson queries. In Ayers et al.'s study, ChatGPT's answers to the AskDocs questions were markedly more detailed and empathetic than those of the doctors, showcasing the utility of such data for benchmarking patient-facing dialogue quality.

### III. METHODOLOGY

#### A. Generation of Synthetic Noise Diagnostic Benchmark

We create synthetic patient self-description narratives by sampling symptoms from the symptom-based disease labelling dataset (SDPD) (Tucker, 2024). For each symptom set, an LLM generates three parallel text segments with varying "noise" levels, labelled as "no-noise," "medium noise" (80–220 words), and "heavy noise" (150–390 words). In total, we produce 3600 narrative triplet examples. Table 1 shows a sample data point.

| | |
|---|---|
| **No Noise** | "My skin is extremely sensitive and quickly irritated by changes in temperature or humidity. My nails have developed dents. I am worried about this sudden change." |
| **Medium Noise** | "Oh dear, my skin… it just gets so red and tender now. Cannot even go outside without gettin' all irritated. Temperature changes, humidity… I swear, I'm like a leaf in the wind. Reminds me of that time at the beach when I was a youngin', got sunburned to a crisp on my poor face… (chuckles) Oh, where's my mind? Ah yes, the skin. And now these dents on my nails! Like little holes or somethin'. Never had that before. Started last month, maybe? Or was it two months ago? Time just gets away from you, doesn't it? Anyway… I'm worried about this. It's like my body's fallin' apart on me." |
| **Heavy Noise** | "Um, doc… I've been having this… this trouble with my skin, you see. It's just so sensitive all of a sudden. Even when I'm out in the garden on a lovely day, and it's warm… oh dear, what was I saying? Ah yes, warm weather. My skin gets all red and irritated, like a rash or something. And then it starts to itch, terrible itching. Can't even wear my nice new scarf without breaking out in hives… And the nails! Goodness gracious, have you ever seen anything like this? Dents on my fingernails, all uneven and whatnot. It's as if… I think it might be that time of year again, you know, when the pollen counts are high… Did I tell you about the time I got a new cat, Whiskers? She was such a sweet thing, always purring on my lap… Where was I? Ah yes, the nails. It's like they're trying to tell me something, but what, I don't know. And it started so suddenly! One day fine, next day… Oh dear, I remember now: it happened after that big storm we had last month. Maybe it was something in the air? Or maybe it's just my imagination playing tricks on me… I'm worried about this sudden change, doc… If I can just get to the bottom of what's causing it all… " |

**Table 1**: Three parallel narratives corresponding to psoriasis diagnosis

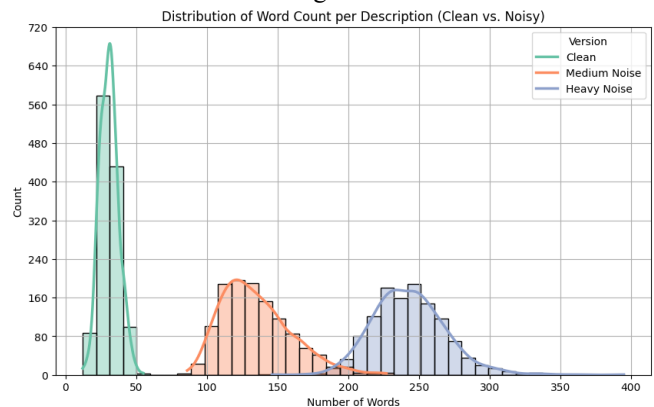

*Figure 1: Word count per noise level*

## B. Benchmarked Diagnostic Models

In our experiments, we evaluate diagnostic performance using a supervised classification setup focused exclusively on fine-tuned models. Specifically, we fine-tune BERT, ClinicalBERT, and Flan-T5 on a clinically clean subset of our dataset, training each model on labeled examples from this curated portion of the data. The primary goal is to measure how these models' diagnostic accuracy degrades when applied to the noisier patient-generated texts in the benchmark, relative to their performance on clean clinical input.

## IV. RESULTS

The supervised evaluation highlights a consistent drop in diagnostic accuracy as input text becomes noisier (Table 2). When trained on clean clinical text, BERT achieved 98.3% accuracy but declined to 86.7% on medium-noise inputs and 79.2% on high-noise inputs. ClinicalBERT followed a similar trend, moving from 97.9% on clean text to 83.8% and 86.2% under medium and high noise, respectively. Flan-T5 demonstrated the most excellent robustness, with only a modest decrease from 97.1% on clean text to 92.5% and 87.1% as noise increased. These results confirm that all fine-tuned models experience measurable degradation when exposed to noisy, patient-generated narratives, and they underscore the relative resilience of Flan-T5 compared to the BERT-based architectures.

| Model | Clean text accuracy | Medium-noise accuracy | High-noise Accuracy |
|---|---|---|---|
| BERT | **98.3%** | 86.7% | 79.2% |
| ClinicalBERT | 97.9% | 83.8% | 86.2% |
| Flan-T5 | 97.1% | **92.5%** | **87.1%** |

**Table 2**: Accuracy of benchmarked models on our Noisy Diagnostic Benchmark

## V. CONCLUSIONS AND FUTURE RESEARCH

This study examines the diagnostic capabilities of large language models when interpreting noisy, free-form patient narratives. We introduced the **Noisy Diagnostic Benchmark (NDB)**, a synthetic dataset designed to emulate varying levels and types of linguistic noise in patient self-descriptions, enabling controlled and reproducible assessment of model robustness under degraded input conditions.

Using NDB, we conducted a **supervised classification** evaluation limited to three fine-tuned models: BERT, ClinicalBERT, and Flan-T5, each trained on a clinically clean subset of the data and then tested on progressively noisier patient-generated texts. The results (Table 2) show apparent degradation from clean to noisy inputs, with Flan-T5 exhibiting the best robustness, followed by BERT and ClinicalBERT.

By releasing NDB and reporting comparative performance across these fine-tuned architectures, this work establishes a reproducible framework for benchmarking diagnostic robustness. It highlights the need for noise-aware training and evaluation protocols in medical NLP. Future work will expand the benchmark with additional clinical scenarios, finer-grained noise categories, and broader linguistic variation. It will explore a wider range of diagnostic models to further characterize the trade-offs between accuracy, robustness, and interpretability in noisy diagnostic settings.